\documentclass[11pt, a4paper]{article}

\usepackage[T1]{fontenc}
\usepackage[utf8]{inputenc} 
\usepackage{graphicx,color}
\usepackage{epsfig}
\usepackage{times} 
\usepackage{amsmath}
\usepackage{amssymb}
\usepackage{caption}
\usepackage{subcaption}
\usepackage{stackrel}

\usepackage{url}
\usepackage{multirow}
\usepackage{cite}

\usepackage{algorithm}
\usepackage{pgf}
\usepackage{pgfplots}
\usepgfplotslibrary{fillbetween}
\usepackage{tikz}
\usetikzlibrary{arrows,automata}
\usetikzlibrary{intersections}
\usetikzlibrary{calc}
\usetikzlibrary{shapes}
\usetikzlibrary{positioning}
\usetikzlibrary{decorations.text}
\pgfdeclarelayer{background}
\pgfdeclarelayer{foreground}
\pgfsetlayers{background,main,foreground}
\usetikzlibrary{calc,patterns,decorations.pathmorphing,decorations.markings}

\newcommand{\proof}{\noindent{\em Proof}.~}
\newcommand{\QED}{$\Box$}

\graphicspath{{figures/}}

\usepackage{enumitem}

\newcommand{\umin}{u_{\rm min}}
\newcommand{\umax}{u_{\rm max}}

\newcommand{\NN}{{\mathcal N}}

\newtheorem{lemma}{Lemma}
\newtheorem{theorem}{Theorem}

\newtheorem{remark}{Remark}
\newcommand{\rr}{\mathbb{R}}
\newcommand{\st}{\mathop{\rm s.t.}\nolimits}
\newcommand{\eqdef}{\triangleq}

\newcommand{\beqar}{\begin{eqnarray}}
\newcommand{\eeqar}{\end{eqnarray}}
\newcommand{\beqarno}{\begin{eqnarray*}}
\newcommand{\eeqarno}{\end{eqnarray*}}
\newcommand{\ba}[1]{\begin{array}{#1}}
\newcommand{\ea}{\end{array}}

\newcommand{\sign}{\mathop{\rm sign}\nolimits}

\newcommand{\matrice}[2]{\left[\hspace*{-.1cm}\ba{#1} #2 \ea\hspace*{-.1cm}\right]}
\newcommand{\smallmat}[1]{\left[ \begin{smallmatrix}#1 \end{smallmatrix} \right]}

\newcommand{\tx}{{\theta_x}}
\newcommand{\ty}{{\theta_y}}

\begin{document}

	\title{Recurrent Neural Network Training 
     with Convex Loss and Regularization Functions by Extended Kalman Filtering}
	
	\author{Alberto Bemporad\thanks{The author is with the IMT School for Advanced Studies, Piazza San Francesco 19, Lucca, Italy. Email: \texttt{\scriptsize alberto.bemporad@imtlucca.it}}}

\maketitle
\thispagestyle{empty}
	
\begin{abstract}
This paper investigates the use of extended Kalman filtering to train recurrent
neural networks with rather general convex loss functions and regularization terms on the network 
parameters, including $\ell_1$-regularization. We show that the learning method
is competitive with respect to stochastic gradient descent
in a nonlinear system identification benchmark and in training a linear system
with binary outputs. We also explore the use of the algorithm in
data-driven nonlinear model predictive control and its relation with disturbance models
for offset-free closed-loop tracking.

{\renewcommand*\@makefnmark{}
\footnotetext{This paper was partially supported by the Italian Ministry of University and Research under the PRIN'17 project ``Data-driven learning of constrained control systems'', contract no. 2017J89ARP.}
\makeatother} 
\end{abstract}


\section{Introduction}
The use of artificial neural networks (NNs) for control-oriented
modeling of dynamical systems, already popular in the nineties \cite{SVD95}, is flourishing again thanks to the wide success
of machine learning in many application domains and to the availability of excellent software
libraries for training NN models. Most frequently, \emph{feedforward} NNs are used for modeling the output function in a neural-network autoregressive model with exogenous inputs (NNARX).
On the other hand, \emph{recurrent} neural networks
(RNNs), as they are state-space models, are often more adequate for capturing the behavior of dynamical systems in a compact way. 
However, contrarily to training NNARX models based on minimizing the one-step-ahead output prediction error, training RNNs is more difficult due to the presence of the hidden states of the network.
To circumvent this issue, procedures for learning
neural state-space models were proposed in~\cite{PB03,MB21}, based
on the idea of finding a (reduced-order) state-space realization of a NNARX model 
during training and considering the values of a (thin) inner layer of the model as the state vector. 

To avoid unrolling the open-loop prediction of an RNN over the entire training dataset,
suboptimal approaches were proposed, such as truncated backpropagation through time (truncated BPTT)~\cite{WP90}. Motivated by the fact that a similar issue of recurrence occurs in minimizing the \emph{simulation} error when training NNARX models, the authors in~\cite{FP20} 
proposed to learn RNNs based on splitting the dataset into smaller batches and penalizing the inconsistency between state predictions across consecutive batches. 
Due to the difficulty of computing gradients of full unrolls and/or
the presence of a large number of optimization variables, these approaches are
used for offline learning and rely on (batch, mini-batch, or stochastic)
gradient descent methods 
that, although often proved to converge only for convex problems,
are widely adopted with success. 

To circumvent their very slow convergence and, at the same time, be able to learn NN models
incrementally when new data become available, training methods based
on extended Kalman filtering (EKF) have been explored in~\cite{SW89} for \emph{feedforward}
networks, treating the weight/bias terms of the network as constant states
to estimate. When dealing with \emph{recurrent} networks, the authors in~\cite{PF94} distinguish between \emph{parallel}-EKF, which estimates both the hidden states and the weights/bias
terms, and \emph{parameter-based} EKF, that only estimates the network parameters, and focus on
the latter approach, whose convergence properties were investigated in~\cite{WH11}.

Parallel-EKF was also studied in~\cite{Wil92} for coinciding
output and state vectors, which is therefore measurable as for NNARX models.
In the context of nonlinear filtering, the authors in~\cite{MM90b} investigated a parallel-EKF approach for the special case of RNNs that are linear in the input and the output. The method was extended
in~\cite{LFB92} to make the EKF implementation more efficient by taking into account how much each network parameter affects the predicted output. An ensemble Kalman filter method was proposed in \cite{MN10}. An important aspect of EKF-based learning methods is that, as remarked in~\cite{IST92}, the filter approximately provides the minimum variance estimate of the model parameters, and hence a
quantification of model uncertainty via the covariance matrix associated with the 
state estimation error. All the aforementioned EKF-based methods consider quadratic penalties on the output prediction errors and the weight/bias terms of the RNN.

In this paper, we also consider a parallel-EKF approach
to recursively learn a general class of RNNs whose state-update and output functions are
described by neural networks, including long-short term memory (LSTM) models~\cite{HS97}, 
under arbitrary convex and twice-differentiable loss functions 
for penalizing output open-loop prediction errors and for regularizing
the weight/bias terms of the RNN, and also extend to the case of $\ell_1$-regularization
for network topology selection. We also compare EKF to different
formulations based on stochastic gradient descent (SGD) methods, in which
only the initial state or also the intermediate states of the RNN are treated as optimization variables.
We show the superiority of EKF with respect to SGD in a 
nonlinear identification benchmark, and apply the method in identifying a linear
dynamical system with binary outputs. We also explore the use of EKF-based
learning for data-driven nonlinear MPC design, showing a relation between the use of constant disturbance models and the adaptation of the bias terms
in the neural networks for offset-free tracking, which we illustrate in a nonlinear control
benchmark problem.

\subsection{Notation}
Given a vector $v\in\rr^n$ and a matrix $A\in\rr^{m\times n}$, 
$v_i$ denotes the $i$th real-valued component of $v$, $A_{i,:}$ the $i$th row of $A$,
$A_{:,j}$ its $j$th column, $A_{ij}$ its $(i,j)$th entry. Given $v\in\rr^n$ and a 
symmetric positive semidefinite matrix $Q=Q'\succeq 0$, $Q\in\rr^{n\times n}$,
we denote by $\|v\|_Q^2$ the quadratic form $v'Qv$, by $\|v\|_1=\sum_{i=1}^n|v_i|$
the 1-norm of $v$, and by $\sign(v)$ the vector whose $i$th component is the sign of $v_i$.
Given $a,b\in\mathbb{N}$, $\delta_{a,b}$ denotes the Kronecker delta function ($\delta_{a,b}=1$ if $a=b$ or $0$ otherwise).

\section{Recurrent neural network model}
\label{sec:RNN}
We consider recurrent neural network (RNN) models with input $u\in\rr^{n_u}$, predicted output $\hat y\in\rr^{n_y}$, and state vector $x\in\rr^{n_x}$ 
whose state-update and output equations are 
described by the following parametric model
\begin{equation}
    \ba{rcl}
        x(k+1)&=&f_x(x(k),u(k),\tx)\\
        \hat y(k)&=&f_y(x(k),u(k),\ty)
    \ea
\label{eq:generic-RNN}
\end{equation}
where $\tx\in\rr^{n_\tx}$ and $\ty\in\rr^{n_\ty}$ collect the parameters 
to learn from data. Special cases of~\eqref{eq:generic-RNN} are recurrent (deep) neural
networks (RNNs)
\begin{subequations}
\begin{equation}
\left\{\ba{rcl}
    v_{1}^x(k)&=&A_{1}^x\smallmat{x(k)\\u(k)}+b_1^x\\
    v_{2}^x(k)&=&A_{2}^x f_1^x(v_{1}^x(k))+b_{2}^x\\
    \vdots& &\vdots\\
    v_{L_x}^x(k)&=&A_{L_x} f_{L_x-1}^x(v_{L_x-1}^x(k))+b_{L_x}^x\\
    x(k+1)&=&v_{L_x}^x(k)
\ea\right.
\label{eq:RNN-x}
\end{equation}
\begin{equation}
\left\{\ba{rcl}
    v_{1}^y(k)&=&A_1^y\smallmat{x(k)\\u(k)}+b_1^y\\
    v_{2}^y(k)&=&A_{2}^y f_1^y(v_{1}^y(k))+b_{2}^y\\
    \vdots& &\vdots\\
    v_{L_y}^y(k)&=&A_{L_y}^y f_{L_y-1}^y(v_{L_y-1}^y(k))+b_{L_y}^y \\
    \hat y(k)&=&f_{L_y}^y(v_{L_y}^y(k))
\ea\right.
\label{eq:RNN-y}
\end{equation}
\label{eq:RNN}%
\end{subequations}
where $L_x-1\geq 0$ and $L_y-1\geq 0$ are the number of hidden layers of the state-update 
and output functions, respectively, $v_i^x\in\rr^{n_i^x}$, $i=1,\ldots L_x-1$,
$v_{L_x}^x\in\rr^{n_x}$,
and $v_i^y\in\rr^{n_i^y}$, $i=1,\ldots L_y$, the values
of the corresponding inner layers,
$f_i^x:\rr^{n_i^x}\to \rr^{n_{i+1}^x}$, $i=1,\ldots L_x-1$,
$f_i^y:\rr^{n_i^y}\to \rr^{n_{i+1}^y}$, $i=1,\ldots L_y-1$
the corresponding activation functions, and $f_{L_y}^y:\rr^{n_{L_y}^y}\to \rr^{n_y}$
the output function, e.g., $f_{L_y}^y(v_{L_y}^y)=v_{L_y}^y$ to model
numerical outputs or $[f_{L_y}^y(v_{L_y}^y)]_i=\left(1+e^{[v_{L_y}^y]_i}\right)^{-1}$,
$i=1,\ldots,n_y$, for binary outputs. The strict causality of the RNN~\eqref{eq:RNN} 
can be simply imposed by zeroing the last $n_u$ columns of $A_1^y$. 

The hyperparameters describing the RNN~\eqref{eq:RNN} are $n_x$, $\{n_i^x\}$, $\{n_y^y\}$,
$\{f_i^x\}$, $\{f_i^y\}$ and dictate the chosen model structure. The parameters
to learn are $A_{1}^x$, $\ldots$, $A_{L_x-1}^x$, $b_{1}^x$, $\ldots$, $b_{L_x-1}^x$,
$A_{1}^y$, $\ldots$, $A_{L_y-1}^y$, $b_{1}^y$, $\ldots$, $b_{L_y-1}^y$, where
$A_{i}^x\in\rr^{n_i^x\times n_{i-1}^x}$ is the matrix of weights for layer $\#i$
of the state-update neural function, 
$b_i^x\in\rr^{n_i^x}$ the corresponding vector of bias terms, and $n_0^x\eqdef n_x+n_u$, $n_L^x\eqdef n_x$,
and $A_{i}^y\in\rr^{n_i^y\times n_{i-1}^y}$ is the matrix of weights for layer $\#i$
of the output neural function, 
$b_i^x\in\rr^{n_i^x}$ the corresponding vector of bias terms, 
and $n_0^y\eqdef n_x+n_u$. In this case, 
$\tx$ and $\ty$ are the vectors obtained
by stacking all the entries of the weight/bias parameters
defining the state and output equations, respectively, with
$n_\tx=\sum_{i=1}^{L_x}n_i^x(n_{i-1}^x+1)$ and $n_\ty=\sum_{i=1}^{L_y}n_i^y(n_{i-1}^y+1)$.

We remark that all the results reported in this paper also apply to \emph{feedforward} neural networks, a special case of~\eqref{eq:RNN} for $n_x=0$. Moreover,
they can be clearly applied also to identify linear models, that is another special
case of~\eqref{eq:RNN} when $L_x=L_y=1$, $b_1^x=0$, $b_1^y=0$, as previously shown in~\cite{Lju79}. 
Another popular instance of model~\eqref{eq:generic-RNN} is the single-layer 
LSTM model~\cite{HS97}. We will consider the form proposed in~\cite{BTFS20}
with output equation as in~\eqref{eq:RNN-y}.

\section{RNN training problem}
\label{sec:learning}
Consider the following RNN learning problem: Given a training dataset $D_N\eqdef\{u(0),y(0),\ldots,u(N-1),y(N-1)\}$, determine an optimal solution $(x_0^*,\theta^*)$
solving the following mathematical program
\begin{equation}
    \ba{rl}
    \stackbin[x_0,\theta]{}{\min} &V(x_0,\theta)\eqdef\displaystyle{r_\theta(\theta)+r_{x}(x_0)+\frac{1}{N}\sum_{k=0}^{N-1}\ell(y(k),\hat y(k))}\\
    \st& \mbox{model equations~\eqref{eq:generic-RNN} with $x(0)=x_0$}
    \ea
\label{eq:min-loss}
\end{equation}
where $\theta\eqdef\smallmat{\tx\\\ty}$,
$\ell:\rr^{n_y}\times \rr^{n_y}\to\rr$ is a loss function
penalizing the dissimilarity between the training samples $y(k)$
and the predicted values $\hat y(k)$ generated by simulating~\eqref{eq:RNN},
and $r_\theta:\rr^{n_\theta}\to\rr$,  $r_x:\rr^{n_x}\to\rr$
are regularization functions. In the examples reported in Section~\ref{sec:experiments} 
we will consider the (weighted) mean-square error (MSE)
loss $\ell_{\rm MSE}(y,\hat y)=\frac{1}{2}\|y-\hat y\|_{W_y}^2$
where $W_y=W_y'\succ 0$ is a weight matrix, 
the (modified) cross-entropy loss
$\ell_{\rm CE\epsilon}(y(k),\hat y)=\sum_{i=1}^{n_y} -y_i(k)\log(\epsilon+\hat y_i)-(1-y_i(k))\log(1+\epsilon-\hat y_i)$ for binary outputs,
the Tikhonov (or $\ell_2$) regularization terms
$r_\theta(\theta)=\frac{\rho_\theta}{2}\|\theta\|_2^2$, $r_x(x_0)=\frac{\rho_x}{2}\|x_0\|_2^2$
with $\rho_\theta,\rho_x\geq 0$, and the $\ell_1$-regularization $r_\theta(\theta)=\lambda\|\theta\|_1$. 

In this paper we consider learning problems based on a single training
dataset $D_N$. The extension to multiple training datasets $D_{N_1}^1,\ldots,D_{N_{n_d}}^{n_d}$,
$D_{N_d}^{n_d}$ $\eqdef$ $\{u^d(0)$, $y^d(0)$, $\ldots$, $u^d(N_d-1),y^d(N_d-1)\}$, $d=1,\ldots,n_d$,
$n_d>1$ can be simply formulated as
\begin{equation}
    \ba{rl}
    \stackbin[x_0^1,\ldots,x_0^{n_d},\theta]{}{\min} &\displaystyle{r_\theta(\theta)+\sum_{d=1}^{n_d}
    r_{x}(x_0^d)+\frac{1}{N_d}\sum_{k=0}^{N_d-1}\ell(y^d(k),\hat y^d(k))}\\
    \st& \hat y^d(k),x^d(k),\ \mbox{and}\ u^d(k)\ \mbox{satisfy model~\eqref{eq:generic-RNN}}\\
    &x^d(0)=x_0^d,\ d=1,\ldots,n_d.
    \ea
\label{eq:min-loss-multiple}
\end{equation}

\subsection{Condensed learning}
\label{sec:condensed}
An optimizer $(x_0^*,\theta_0^*)$ of Problem~\eqref{eq:min-loss} can be computed
by different unconstrained nonlinear programming (NLP) solvers~\cite{NW06},
including the approach recently proposed in~\cite{Bem22}. 
For a given value of $(x_0,\theta)$, evaluating $V(x_0,\theta)$  
requires processing an \emph{epoch}, i.e., the entire training dataset.
Since the cost function in~\eqref{eq:min-loss} is not separable
due to the dynamic constraints~\eqref{eq:RNN}, a 
gradient descent method corresponds to the steepest-descent steps
\begin{equation}
    \matrice{c}{x_0^{t+1}\\\theta^{t+1}} = \matrice{c}{x_0^{t}\\\theta^{t}}-\alpha_t
    \matrice{c}{\frac{\partial V}{\partial x_0}(x_0^t,\theta^t)\\[.3em]
    \frac{\partial V}{\partial \theta}(x_0^t,\theta^t)}
\label{eq:batch-SGD}
\end{equation}
where $\alpha_t$ is the learning rate at epoch $t=1,\ldots,N_e$.
By applying the chain rule for
computing derivatives, the gradient of the objective function with respect to $(x_0,\theta)$
can be evaluated efficiently by BPTT~\cite{Wer90}.
However, gradient computation involves well-known issues, for example
in the case of vanilla RNNs \emph{vanishing gradients}~\cite{Hoc98} 
and \emph{exploding gradients} 
effects have been reported.
While heuristic remedies to alleviate this effect,
such as approximating the gradients by truncated BPTT or different architectures such as LSTMs~\cite{HS97} were proposed, the root-cause
lies in having \emph{condensed} the problem by eliminating the intermediate
variables $x(k)$, $k=1,\ldots,N-1$. 

Problem~\eqref{eq:min-loss} can be interpreted as a finite-horizon
optimal control problem with free initial state $x_0$ (penalized by $r_x(x_0)$),
where $\theta(k)\equiv \theta$ is the vector of manipulated
inputs to optimize, $r_\theta(\theta)$ the corresponding penalty, $\hat y(k)$
the controlled output, $y(k)$ the
output reference, $u(k)$ a measured disturbance (alternatively,
$x_0$ could be also seen as an input only acting at time $-1$ to change
$x(0)$ from $x(-1)=0$).
As well known in solving MPC problems,
the effect of \emph{condensing} the problem 
(a.k.a. ``direct single shooting''~\cite{HR71}) by eliminating state variables
potentially leads to numerical
difficulties. For example in standard linear MPC
formulations an unstable linear prediction model
$x(k+1)$ = $Ax(k)$ + $Bu(k)$ leads to a quadratic program with ill-conditioned Hessian
matrix, due to the presence of the terms $A^k$ that appear in constructing
the program (cf.~\cite{BC21}). The reader is also referred to the recent work~\cite{ELTSP21}
for a re-interpretation of the training problem of feedforward neural networks as an optimal 
control problem and the benefits of avoiding condensing from a solution-algorithm perspective.

\subsection{Relaxed non-condensed learning}
\label{sec:non-condensed}
By following the analogy with MPC, the problem in~\eqref{eq:min-loss}
can be solved in \emph{non-condensed} form 
(a.k.a. ``direct multiple shooting''~\cite{BP84})
by also treating $x(1)$, $\ldots$, $x(N-1)$
(and possibly also $v_i^x(k)$, $v_j^y(k)$) as additional optimization variables,
subject to the equality constraints imposed by the RNN model equations~\eqref{eq:RNN}.
By further relaxing the model equations we get the following unconstrained nonlinear optimization
problem
\beqar
\stackbin[{\scriptsize\ba{c}x_0,\theta\\x_1,\ldots,x_{N-1}\ea}]{}{\min}\hspace*{-1.2cm} &&r_\theta(\theta)+r_x(x_0)+\frac{1}{N}\sum_{k=0}^{N-1}\ell(y(k),
f_y(x_k,u(k),\ty))\nonumber\\[-.5em]&&+\frac{\gamma}{2N}\sum_{k=0}^{N-2}\|x_{k+1}-f_x(x_k,u(k),\tx)\|_2^2
\label{eq:min-loss-relaxed}
\eeqar
where $\gamma>0$ is a scalar penalty
promoting the consistency of the state sequence with 
the RNN model equations~\eqref{eq:RNN-x}.

Note that in the case of MSE loss and $\ell_2$-regularization,~\eqref{eq:min-loss-relaxed} can be solved as a nonlinear least-squares problem, for which efficient solution methods exist~\cite{NW06}.
The relaxation~\eqref{eq:min-loss-relaxed}
is also amenable for stochastic gradient-descent iterations
that only update $(x_k,x_{k+1},\theta)$ when sample $k$ is processed:
\begin{equation}
    \ba{l}
    \displaystyle{\matrice{c}{x_{k+1}^{k+1}\\x_{k}^{k+1}\\\theta^{k+1}} = 
    \matrice{c}{x_{k+1}^{k}\\x_{k}^{k}\\\theta^{k}} -
    \alpha_k\smallmat{0\\\frac{\partial \ell}{\partial x}(y(k),f_y(x_k^k,u(k),\theta_y^k))\\[.3em]
    \frac{\partial \ell}{\partial \theta}(y(k),f_y(x_k^k,u(k),\theta_y^k))}}\\[2em]
    \displaystyle{-\gamma\alpha_k
    \smallmat{I\\
\frac{\partial f_x}{\partial x}(x_k^k,u(k),\theta_x^k)'\\[.3em]
\frac{\partial f_x}{\partial \theta}(x_k^k,u(k),\theta_x^k)'}(x_{k+1}-f_k)-\frac{\alpha_k}{N}\smallmat{0\\\nabla r_x(x_0^k)\delta_{k,0}\\\nabla r_\theta(\theta^k)}}
\ea
\label{eq:SGD-relaxed}
\end{equation}
where $f_k\eqdef f_x(x_k^k,u(k),\theta_x^k))$, $k=0,\ldots,N-1$. Clearly,~\eqref{eq:SGD-relaxed}
can be also run on multiple epochs by resetting $k=0$ at each epoch $t=1,\ldots,N_e$.

Finally, we remark that in case of multiple training datasets ($n_d>1$),
the number of optimization variables remains $Nn_x+n_\theta$, where
$N=\sum_{d=1}^{n_d}N_d$, and simply some regularization terms $\gamma\|x_{k+1}-f_x(x_k,u_k,\tx)\|_2^2$ are skipped in~\eqref{eq:min-loss-relaxed} to take into account that
the initial state of a new training sequence is not related to the final predicted state
of the previous sequence.

\subsection{Relaxed partially-condensed learning}
\label{sec:partially-condensed}
The non-condensed form~\eqref{eq:min-loss-relaxed}
has $n_\theta+Nn_x$ optimization variables, 
where in the present context of control-oriented RNN models usually $Nn_x\gg n_\theta$.
\emph{Partial condensing}~\cite{Axe15} can reduce the training problem as follows. Let us split the dataset $D_N$ into $M$ batches, $M\leq N$, of lengths $L_1,\ldots,L_M$, respectively (for example, $L_1=L_2=L_{M-1}=\lceil\frac{N}{M}\rceil$,
$L_M=N-(M-1)\lceil\frac{N}{M}\rceil$).
Then, we solve the following problem:
\begin{equation}
    \ba{rl}
    \stackbin[x_0,x_1,\ldots,x_{M-1},\theta]{}{\min}\hspace*{-.2cm} &
\displaystyle{\frac{1}{N}
    \sum_{j=0}^{M-1}\sum_{h=0}^{L_j-1}\ell(y(k_{ij}),f_y(\hat x_{h|j},u(k_{ij}),\ty)}
\\&\hspace{-2cm}
    +r_\theta(\theta)+r_x(x_0)+\displaystyle{\gamma\frac{N-1}{2N(M-1)}\sum_{j=0}^{M-2}\|x_{j+1}-\hat x_{L_j|j}\|_2^2}\vspace*{-.2cm}
    \ea
\label{eq:min-loss-partial}
\end{equation}
where $k_{ij}\eqdef h+\sum_{s=0}^{j-1}L_s$ and
$\hat x_{h|j}$ is the hidden state predicted by iterating~\eqref{eq:RNN-x}
over $h$ steps from the initial condition $x_j$ under the input excitation $u(L_j)$,
$\ldots$, $u(L_j+h-1)$,
$h=1,\ldots,L_j$, $j=0,\ldots,M-1$. Similarly
to the approach of~\cite{FP20}, problem~\eqref{eq:min-loss-partial}
can be solved by an SGD method by processing $L_j$ samples at the time,
that results in updating $(x_{j+1},x_j,\theta)$ at each SGD iteration.
Note that~\eqref{eq:min-loss-partial} includes~\eqref{eq:min-loss-relaxed}
as a special case by setting $M=N$, $L_j\equiv 1$, and~\eqref{eq:min-loss} for $M=1$, $L_1=N$.

\section{Training by Extended Kalman filtering}
To address both the \emph{recursive} estimation of $\theta$ from 
real-time streams of input and output measurements and counteract
the slow convergence of SGD methods, we consider
the use of extended Kalman filtering (EKF) techniques as in~\cite{MM90b,LFB92} to recursively update $\theta$ and the current hidden state vector. To this end, let us rewrite model~\eqref{eq:RNN} 
as the following nonlinear system affected by noise
\begin{equation}
\ba{rcl}
    x(k+1)&=&f_x(x(k),u(k),\theta_x(k))+\xi(k)\\[.5em]    
    y(k)&=&f_y(x(k),u(k),\theta_y(k))+\zeta(k)\\[.5em]
    \theta(k+1)&=&\theta(k)+\eta(k),\quad \theta(k)\eqdef\smallmat{\tx(k)\\\ty(k)}
\ea
\label{eq:EKF-model}
\end{equation}
where $\xi(k)\in\rr^{n_x}$, $\zeta(k)\in\rr^n_y$,
and $\eta(k)\in\rr^{n_\theta}$ are white, zero-mean, noise
vectors with covariance matrices $Q_x(k)$, $Q_y(k)$, and $Q_\theta(k)$, respectively,
with $Q_x(k)=Q_x(k)'\succeq 0$, $Q_x(k)\in\rr^{n_x\times n_x}$, $Q_y(k)=Q_y(k)'\succ 0$,
$Q_y(k)\in\rr^{n_y\times n_y}$, $Q_\theta(k)=Q_\theta(k)'\succeq 0$, $Q_\theta(k)\in\rr^{n_\theta\times n_\theta}$ for all $k$. 

The model coefficients $\hat \theta(k)$ and the hidden state $\hat x(k)$ are 
estimated with data up to time $k$ according to the following classical  
recursive EKF updates:
\begin{subequations}
\begin{equation}
\ba{rcl}
C(k)&\hspace*{-1em}=&\hspace*{-1em}\left.\matrice{ccc}{\frac{\partial f_y}{\partial x}&0& \frac{\partial f_y}{\partial \theta_y}}\right|_{\hat\theta(k|k-1),\hat{x}(k|k-1),u(k)}\\[1em]
M(k)&\hspace*{-1em}=&\hspace*{-1em}P(k|k-1)C(k)'[C(k)P(k|k-1)C(k)'\\&&+Q_y(k)]^{-1}\\
e(k)&\hspace*{-1em}=&\hspace*{-1em}y(k)-f_y(\hat{x}(k|k-1),u(k),\hat\theta_y(k|k-1))\\[.5em]
\matrice{c}{\hat{x}(k|k)\\\hat{\theta}(k|k)}&=&
\matrice{c}{\hat{x}(k|k-1)\\\hat{\theta}(k|k-1)}+M(k)e(k)\\[1em]
P(k|k)&\hspace*{-1em}=&\hspace*{-1em}(I-M(k)C(k))P(k|k-1)
\ea
\label{eq:EKF-full-time}
\end{equation}
\begin{equation}
{\small \ba{rcl}
\matrice{c}{\hat{x}(k+1|k)\\\hat\theta(k+1|k)}&=&\matrice{c}{f_x(\hat{x}(k|k),u(k),\hat{\theta}_x(k|k))\\\hat\theta(k|k)}\\[.5em]
A(k)&\hspace*{-1em}=&\hspace*{-1em}\left.\matrice{ccc}{\frac{\partial f_x}{\partial x} & 
\frac{\partial f_x}{\partial \theta_x}&0\\[.2em]0& I & 0\\[.2em]0 &0 & I}\right|_{\hat\theta(k|k),\hat{x}(k|k),u(k)}\\[1em]
P(k+1|k)&\hspace*{-1em}=&\hspace*{-1em}A(k)P(k|k)A(k)'+\smallmat{Q_x(k)&0\\0&Q_\theta(k)}.
\ea}
\end{equation}
\label{eq:EKF-full}%
\end{subequations}
Note that in the single output case ($n_y=1$), the matrix inversion in~\eqref{eq:EKF-full-time}
becomes a simple division. 

Let us recall the equivalence between the EKF~\eqref{eq:EKF-full} and Newton's method~\cite[Sect. 5.2]{HRW12} for solving~\eqref{eq:min-loss-relaxed} in the case of loss $\ell_{\rm MSE}$ and $\ell_2$-regularization, and of an additional penalty on updating $\theta$:
\begin{equation}
\ba{rl}
\stackbin[{\scriptsize\ba{c}
x_0,x_1,\ldots,x_{N-1}\\
\theta_0,\theta_1,\ldots,\theta_{N-1}\ea}]{}{\min} &
\frac{1}{2}\left\|\smallmat{x_0\\\theta}-
\smallmat{x(0|-1)\\\theta(0|-1)}\right\|^2_{P(0|-1)^{-1}} 
\\[-0.8em]
&\hspace*{-1.4em}
+\frac{1}{2}\sum_{k=0}^{N-1}\|y(k)-
f_y(x_k,u(k),\theta_{yk})\|^2_{Q_y^{-1}(k)}
\\
&\hspace*{-1.4em}
+\frac{1}{2}\sum_{k=1}^{N-2}\|x_{k+1}-f_x(x_k,u(k),\theta_{xk})\|^2_{Q_x^{-1}(k)}\\
&\hspace*{-1.4em}
+\|\theta_{k+1}-\theta_k\|^2_{Q_\theta^{-1}(k)}. 
\ea
\label{eq:EKF-Newton}
\end{equation}
By dividing the cost function in~\eqref{eq:EKF-Newton} by $N$, we have that 
$Q_x=\frac{1}{\gamma}I$, $Q_y=W_y^{-1}$, while $Q_\theta(k)=\frac{1}{\alpha_k}I$ induces a learning-rate due to minimizing the additional term $\frac{\alpha_k}{2}\|\theta_{k+1}-\theta_k\|_2^2$.
The initial vector $\smallmat{x(0)\\\theta(0)}$ is treated as a random vector with mean
$\smallmat{x(0|-1)\\\theta(0|-1)}$ and covariance $P(0|-1)=P(0|-1)'\succeq 0$, $P(0|-1)\in\rr^{(n_x+n_\theta)\times (n_x+n_\theta)}$. By the aforementioned analogy between EKF and Newton's method, we have that 
\begin{equation}
    P(0|-1)=\smallmat{\frac{1}{N\rho_x}I&0\\0&\frac{1}{N\rho_\theta}I}    
\label{eq:P0-1}
\end{equation}
is directly related to the $\ell_2$-regularization terms
$r_\theta(\theta)=\frac{\rho_\theta}{2}\|\theta\|_2^2$, $r_x(x_0)=\frac{\rho_x}{2}\|x_0\|_2^2$
used in~\eqref{eq:min-loss-relaxed}
when $x(0|-1)=0$, $\theta(0|-1)=0$.
We remark that the EKF iterations~\eqref{eq:EKF-full}
are not guaranteed to converge to a global minimum
of the posed training problem. The reader is referred to the EKF convergence 
results reported in~\cite{Lju79,BRD97} for further details.

In the next section, we show how to extend EKF-based training to handle
generic strongly convex and twice differentiable loss functions $\ell$ and regularization terms $r_\theta$, $r_x$.

\subsection{EKF with general output prediction loss}
\begin{lemma}
Let $\ell:\rr^{n_y}\times\rr^{n_y}\to\rr$ be strongly convex and twice differentiable with respect to its second argument $\hat y$. Then by setting
\begin{subequations}
\beqar
    Q_y(k)&\eqdef&\left(\frac{\partial^2 \ell(y(k),\hat y(k|k-1))}{\partial \hat y^2}\right)^{-1}
    \label{eq:EKF-loss-1}\\
    e(k)&=& 
    -Q_y(k)\frac{\partial \ell(y(k),\hat y(k|k-1))}{\partial \hat y}
\label{eq:EKF-loss-2}%
\eeqar
\label{eq:EKF-loss}%
\end{subequations}
\label{lemma:loss}
the EKF updates~\eqref{eq:EKF-full} attempt at minimizing the loss $\ell$.
\end{lemma}
\proof
At each step $k$, let us take a second-order Taylor expansion of $\ell(y(k),\hat y)$ around $\hat y(k|k-1)$ to approximate
\begin{equation}
    \ba{rcl}
\arg\min\ell(y(k),\hat y)&\approx &\arg\min \{\frac{1}{2}\Delta y_k'Q^{-1}_y(k)\Delta y_k\\
&&+\varphi_k'\Delta y_k\}\\
&\hspace*{-12em}=&\hspace*{-6em}\arg\min\{\frac{1}{2}\|\hat y(k|k-1)-Q_y(k)\varphi_k
-\hat y\|_{Q^{-1}_y(k)}^2\}
\ea
\label{eq:ell-approx}
\end{equation}
where $\Delta y_k=\hat y-\hat y(k|k-1)$
and $\varphi_k\eqdef\frac{\partial \ell(y(k),\hat y(k|k-1))}{\partial \hat y}$.

Due to the parallel between EKF and Newton's method recalled in~\eqref{eq:EKF-Newton},
feeding the measured output $y(k)=\hat y(k|k-1)-Q_y(k)\varphi_k$ 
gives $e(k)$ and $Q_y(k)$ as in~\eqref{eq:EKF-loss}.
\hfill\QED

\begin{remark}
Note that for the MSE loss $\ell_{\rm MSE}(y(k),\hat y)=\frac{1}{2}\|y(k)-\hat y\|_{W_y}^2$,
as $Q_y(k)=W_y^{-1}$, 
we get $\varphi_k=Q_y^{-1}(k)(\hat y(k|k-1)-y(k))$ and hence from~\eqref{eq:EKF-loss-2} the classical output prediction error term $e(k)=y(k)-\hat y(k|k-1)$. \hfill\QED
\end{remark}

\begin{remark}
When the modified cross-entropy loss $\ell_{\rm CE\epsilon}(y(k),\hat y)$ 
is used to handle binary outputs $y(k)\in\{0,1\}$ and 
predictors $\hat y(k|k-1)\in[0,1]$ we get 
$\varphi_k=
-\frac{y(k)}{\epsilon+\hat y(k|k-1)}+\frac{1-y(k)}{1+\epsilon-\hat y(k|k-1)}$ and
\begin{subequations}
\begin{equation}
    Q_y(k)=
\left(\frac{y(k)}{(\epsilon+\hat y(k|k-1))^2}+\frac{1-y(k)}{(1+\epsilon-\hat y(k|k-1))^2}\right)^{-1}.
\label{eq:Qy-CEloss}
\end{equation}
Hence, from~\eqref{eq:EKF-loss-2}, we get $e(k)=-1-\epsilon-\hat y(k|k-1)$
for $y(k)=0$ and $e(k)=\epsilon+\hat y(k|k-1)$ for $y(k)=1$,
or equivalently 
\begin{equation}
    e(k)=(1+2\epsilon)y(k)+\hat y(k|k-1)-1-\epsilon.
\label{eq:e-CEloss}
\end{equation}
\label{eq:CEloss}%
\end{subequations}
Note that $\epsilon>0$ is used to avoid possible numerical issues in
computing $Q^{-1}_y(k)$ when $\hat y(k|k-1)$ tends to 0 or 1.
\hfill\QED
\end{remark}

\subsection{EKF with generic regularization terms}
We have seen in~\eqref{eq:P0-1} how a quadratic regularization $r_\theta$
can be embedded in the EKF formulation by properly defining $P(0|-1)$. We now extend
the formulation to handle more general regularization terms.
\begin{lemma}
Consider the generic regularization term
\[
    r_\theta(\theta)=\Psi(\theta)+\frac{1}{2}\rho_\theta\|\theta\|_2^2
\]
and let $\Psi:\rr^{n_\theta}\to\rr$ be strongly convex and twice differentiable. Then by extending
model~\eqref{eq:EKF-model} with the additional system output 
\begin{equation}
y_\Psi(k)=\theta(k)+\mu_\Psi(k)    
\label{eq:yPsi}
\end{equation}
where $\mu_\Psi(k)$ has zero mean and covariance 
\begin{subequations}
\begin{equation}
    Q_\Psi(k)=(\nabla_\theta^2\Psi(\theta(k|k-1)))^{-1}
\label{eq:QPsi-reg}
\end{equation}
and by feeding the error term
\begin{equation}
e_\Psi(k)=-Q_\Psi(k)\nabla_\theta \Psi(\theta(k|k-1))
\label{eq:EKF-reg-2}
\end{equation}
\label{eq:generic-regularization}
\end{subequations}
the EKF updates~\eqref{eq:EKF-full} attempt at minimizing $r_\theta(\theta)$.
\label{lemma:EKF-reg}
\end{lemma}
\proof
Since $\Psi(\theta)=\frac{1}{N}\sum_{k=0}^{N-1}\Psi(\theta)$, at a given prediction step $k$ consider the further loss term $\Psi(\theta)$ and its approximate minimization around $\theta(k|k-1)$
\[
    \ba{rcl}
\arg\min \Psi(\theta)&\approx &\arg\min\left\{\frac{1}{2}\Delta\theta_k' Q^{-1}_\Psi\Delta\theta_k+
\gamma_k'\Delta\theta_k\right\}\\
&\hspace*{-8em}=&\hspace*{-4em}\arg\min\left\{\frac{1}{2}\|\Delta\theta_k+Q_\Psi(k)\gamma_k\|_{Q^{-1}_\Psi(k)}^2\right\}
\ea
\]
where $\Delta\theta_k\eqdef \theta-\theta(k|k-1)$ and
$\gamma_k\eqdef\nabla_\theta \Psi(\theta(k|k-1))$. 
By feeding the measurement $y_\Psi(k)=\theta(k|k-1)-Q_\Psi^{-1}(k)\gamma_k$
to the EKF, as in~\eqref{eq:EKF-loss} we get the additional error term $e_\Psi(k)=-Q_\Psi(k)\gamma_k$
and the corresponding output Jacobian matrix $C_\Psi(k)=[0\ I]$. 
\hfill\QED

Note that for $\Psi(\theta)=\frac{1}{2}\bar\rho_\theta\sum_{i=1}^{n_\theta}\theta_i^2$,
then $Q_\Psi(k)=\frac{1}{\bar \rho_\theta}I$ and $e_\Psi(k)=-\hat\theta(k|k-1)$. 
Next Lemma~\ref{lemma:Psi-sep} specializes the result of Lemma~\ref{lemma:EKF-reg} to
the case of separable regularization functions.

\begin{lemma}
Let $\Psi(\theta)=\sum_{i=1}^{n_\theta}\psi_i(\theta_i)$
and assume that each function $\psi_i:\rr\to\rr$ is
strongly convex and twice differentiable.
Then, after each measurement update $k$
and before performing the time update in~\eqref{eq:EKF-full}, minimizing the additional 
loss $\Psi$ corresponds to ($i$) setting
\begin{subequations}
\begin{equation}
    \ba{lcl}
\matrice{c}{\hat{x}(k_0)\\\hat{\theta}(k_0)}&=&
\matrice{c}{\hat{x}(k|k-1)\\\hat{\theta}(k|k-1)}+M(k)e(k)\\[1em]
    P(k_0)&=&(I-M(k)C(k))P(k|k-1)
\ea
\end{equation}
($ii$) iterating
\begin{equation}
\ba{rcl}
    C_{\Psi i}(k_i)&=&I_{n_x+i,:}\\[.3em] 
    e_{\Psi i}(k_i)&=&-\frac{\psi'_i(\hat\theta_i(k_{i-1}))}{\psi''_i(\hat\theta_i(k_{i-1}))}
    \\[.8em]
    M(k_i)&=&\frac{P_{:,n_x+i}(k_{i-1})}{P_{n_x+i,n_x+i}(k_{i-1})+
    1/\psi''_i(\hat\theta_i(k_{i-1}))}\\[1em]
\matrice{c}{\hat{x}(k_i)\\\hat{\theta}(k_i)}&=&
\matrice{c}{\hat{x}(k_{i-1})\\\hat{\theta}(k_{i-1})}
+M(k_i)e_{\Psi i}(k_i)\\[1em]
    P(k_i)&=&
    P(k_{i-1})-M(k_i)P_{n_x+i,:}(k_{i-1})\\
\ea
\label{eq:EKF-gen-loss-i}
\end{equation}
for $i=1,\ldots,n_\theta$, and ($iii$) assigning
\begin{equation}
\matrice{c}{\hat{x}(k|k)\\\hat{\theta}(k|k)}=
\matrice{c}{\hat{x}(k_{n_\theta})\\\hat{\theta}(k_{n_\theta})},\quad
P(k|k)=P(k_{n_\theta}).
\end{equation}%
\label{eq:EKF-gen-loss}%
\end{subequations}
\label{lemma:Psi-sep}
\end{lemma}
\proof The result simply follows due to the independence of each component $\mu_{\Psi i}(k)$ in~\eqref{eq:yPsi}, $i=1,\ldots,n_\theta$, and $\zeta(k)$ in~\eqref{eq:EKF-model},
see, e.g.,~\cite[Sect. 10.2.1]{Gib11}.
In fact, the multi-output measurement update of the EKF can be processed sequentially, i.e., first $y(k)$ to get $\hat x(k_0)$, $\hat \theta(k_0)$, $P(k_0)$,
and then the outputs $y_{\Psi i}(k)$ one by one, for $i=1,\ldots,n_\theta$,
finally getting $\hat x(k|k)$, $\hat \theta(k|k)$, and $P(k|k)$. 
\hfill\QED

\subsection{EKF-based training with $\ell_1$-regularization}
When training RNN models several degrees of freedom exist in selecting the model structure,
i.e., the number of layers and of neurons in each layer of the feedforward
neural networks $f_x$, $f_y$. It is common to start with a large enough
number of parameters and then promote the sparsity of $\theta$ by introducing
the penalty $\lambda \|\theta\|_1$ on $\theta$. We next show how to handle such a
sparsifier in our EKF-based setting.

\begin{subequations}
\begin{theorem}
Minimizing the additional loss $\lambda\|\theta\|_1$ corresponds
to the measurement updates as in~\eqref{eq:EKF-gen-loss} with~\eqref{eq:EKF-gen-loss-i} replaced by
\begin{equation}
    \ba{rcl}
\hspace*{-.5em}\matrice{c}{\hat{x}(k_i)\\\hat{\theta}(k_i)}&\hspace*{-.3em}=&\hspace*{-.3em}
\matrice{c}{\hat{x}(k_{i-1})\\\hat{\theta}(k_{i-1})}-\lambda \sign(\hat\theta_i(k_{i-1}))P_{:,n_x+i}(k_{i-1})\\[1em]
    P(k_i)&=&P(k_{i-1}),\quad i=1,\ldots,n_\theta
\ea
\label{eq:l1-theta-update-1}
\end{equation}
where $\sign:\rr\to\{-1,0,1\}$ is the standard sign function.
\label{th:ekf-L1}
\end{theorem}
\proof Consider the following smooth version
of the 1-norm 
$\Psi(\theta)$ = $\lambda\sum_{i=1}^{n_\theta}$ $(\theta_i^2+\tau\theta_i^4)^\frac{1}{2}$,
$\tau>0$, and the associated derivatives $\psi_i'(\theta_i)=\lambda
\frac{2\tau\theta_i^3+\theta_i}{(\tau\theta_i^4 + \theta_i^2)^\frac{1}{2}}$,
$\psi''_i(\theta_i)=\lambda\frac{\tau\theta_i^4(2\tau\theta_i^2 + 3)}{(\tau\theta_i^4 + \theta_i^2)^\frac{3}{2}}$.
By applying Lemma~\ref{lemma:Psi-sep} we get the following updates
 $    e_{\Psi i}(k_i)=-\frac{2\tau^2\hat\theta_i^4 + 3\tau\hat\theta_i^2 + 1}{\tau\hat\theta_i(2\tau\hat\theta_i^2 + 3)}$,
 $M(k_i)=\frac{\lambda\tau \hat\theta_i^4(2\tau\hat\theta_i^2+3)P_{:,n_x+i}}
 {\lambda\tau\hat\theta_i^4(2\tau\hat\theta_i^2+3)\pi_{ii}+(\tau\hat\theta_i^4+\hat\theta_i^2)^{\frac{3}{2}}}
 $
where for simplicity we set $\pi_{ii}\eqdef P_{n_x+i,n_x+i}$
and omitted ``$(k_{i-1})$'',
and we also assumed $\theta_i\neq 0$.
For $\tau\rightarrow 0$ we get
$
M(k_i)e_{\Psi i}(k_i)\rightarrow -\frac{\lambda\hat\theta_i^3}{(\hat\theta_i^2)^\frac{3}{2}}P_{:,n_x+i} =-\lambda\sign(\hat\theta_i)P_{:,n_x+i}
$
and $M(k_i)\rightarrow 0$.
For $\hat\theta_i\rightarrow 0$ we get $M(k_i)e_{\Psi i}(k_i)\rightarrow 0$
and $M(k_i)\rightarrow 0$ for all $\tau>0$. As $M(k_i)\rightarrow 0$, we also get 
$P(k_i)=P(k_{i-1})$ in~\eqref{eq:EKF-gen-loss-i}.
\hfill\QED

An alternative to~\eqref{eq:l1-theta-update-1} is to evaluate all the sign terms
upfront at $\hat\theta(k_{0})=\hat\theta(k|k-1)$, leading to the update
\begin{equation}
    \ba{rcl}
    \displaystyle{\matrice{c}{\hat x(k|k)\\\theta(k|k)}}&=&\displaystyle{\matrice{c}{\hat x(k|k-1)\\\theta(k|k-1)}}+M(k)e(k)-\lambda P(k|k-1)\matrice{c}{0\\\sign(\hat\theta(k|k-1))}\\[1em]
    P(k|k)&=&(I-M(k)C(k))P(k|k-1)
    \ea
\label{eq:l1-theta-update-2}
\end{equation}
\label{eq:l1-theta-update}
\end{subequations}

\subsection{Complexity of EKF-based training}
We briefly discuss the numerical complexity of the proposed EKF-based 
training method. The iterations~\eqref{eq:EKF-full} require the following
steps: forming matrix $C(k)$ requires evaluating $n_y(n_x+n_\ty)$ partial derivatives;
evaluating $M(k)$ requires computing $D_1(k)\eqdef P(k|k-1)C(k)'$, $D_2(k)\eqdef C(k)D_1(k)$,
the $n_y\times n_y$ inverse symmetric matrix $D_3(k)\eqdef [Q_y(k)+D_2(k)]^{-1}$,
and then evaluate $M(k)=D_1(k)D_3(k)$, which has an overall complexity 
$O(n_y(n_x+n_\theta)^2+(n_x+n_\theta)n_y^2+n_y^3)$; note that no matrix inversion
is required in~\eqref{eq:EKF-full} in the single-output case ($n_y=1$),
as it becomes a simple division. Evaluating $P(k|k)=P(k|k-1)-M(k)D_1(k)'$
has complexity $O(n_y(n_x+n_\theta)^2)$; forming matrix $A(k)$ requires 
evaluating $n_x(n_x+n_\tx)$ partial derivatives; finally, updating
$P(k+1|k)$ requires $O(n_x^3+n_x(n_\tx^2+n_\ty^2)+n_x^2n_\theta)$ operations.
The reader is referred to~[Ch.~10]\cite{Gib11} for different implementations
of the EKF, such as in factored or square-root form.

In the multi-output case ($n_y>1$), an alternative and slightly different formulation 
to avoid inverting $Q_y(k)+D_2(k)$ is to take $Q_y(k)$ diagonal and treat measurements updates 
one by one as in~\eqref{eq:EKF-gen-loss}. The drawback
of this approach is that $M(k)$ and $P(k|k)$ must be updated $n_y$ times. Since
typically $n_y\ll n_\theta$, in general this would be computationally heavier
than performing the computations in~\eqref{eq:EKF-full}.

When the additional $\ell_1$-penalty $\|\theta\|_1$ is included, Eq.~\eqref{eq:l1-theta-update} requires extra $O(n_\theta(n_x+n_\theta))$ operations. When including a more
general separable penalty $\Psi(\theta)$ $=$ $\sum_{i=1}^{n_\theta}\psi_i(\theta)$,
$Q_\Psi(k)$ is an $n_\theta\times n_\theta$ diagonal matrix and
Eq.~\eqref{eq:generic-regularization} requires computing $n_\theta$ second
derivatives $\frac{d^2\psi_i}{d\theta_i^2}$ and their reciprocals,
and $e_\Psi(k)$ involves $O(n_\theta)$ operations. Then~\eqref{eq:EKF-gen-loss} has a complexity
$O(n_\theta(n_\theta+n_x)^2)$.

We finally remark that, according to our numerical experience, the main computation complexity is due to computing 
the partial derivatives in $A(k)$ and $C(k)$ (e.g., by back-propagation), 
especially in the case of neural networks with a large number $L_x$, $L_y$
of layers. Clearly, the computation of Jacobian
matrices is required by all gradient-based training algorithms.

\subsection{Initial state reconstruction}
Given a model $\theta$ and a dataset
$\{\bar u(0),\bar y(0),\ldots,\bar u(\bar N-1),\bar y(\bar N-1)\}$,
testing the prediction capabilities of the model in open-loop simulation 
requires determining an appropriate initial state $\bar x(0)$.
A way to get $\bar x(0)$ is to solve the following state-reconstruction 
problem with $n_x$ variables
\begin{equation}
    \min_{\bar x_0} r_x(x_0)+\frac{1}{\bar N}\sum_{k=0}^{\bar N-1}\ell(\bar y(k),\hat y(k))
\label{eq:min-loss-state}
\end{equation}
where $\hat y(k)$ are generated by iterating~\eqref{eq:RNN}
from $x(0)=\bar x_0$, or its equivalent non-condensed or partially-condensed form.

Solving~\eqref{eq:min-loss-state} is not only useful to test a trained
model on new data, but also when running the EKF~\eqref{eq:EKF-full} offline
on training data over multiple epochs. In this case, Problem~\eqref{eq:min-loss-state}
can provide a suitable value for $x(0|-1)$ for the new epoch based on the last vector $\hat \theta(N-1|N-1)$ learned, that is used in~\eqref{eq:min-loss-state} 
and as the initial
condition $\theta(0|-1)$ for the new epoch. We remark that when the EKF is run on $N_e$ epochs
and $P(0|-1)$ is set equal to the value $P(N|N-1)$ from the previous epoch, 
in~\eqref{eq:P0-1} one should divide by $N_eN$ and not by $N$ for consistency.

In the case of multiple training experiments $n_d>1$, it is enough to run the EKF
on each dataset $D_{N_d}^{n_d}$ by always resetting the initial state $x(0)^d$
as in~\eqref{eq:min-loss-state} based on the first $\bar N$ samples in $D_{N_d}^{n_d}$,
while vector $\theta$ and the covariance matrix $P$ get propagated across experiments
and epochs.

\section{Nonlinear model predictive control}
In a fully adaptive case, the EKF as in~\eqref{eq:EKF-full} can be used online both to
estimate the state and to update the parameters of the model
from streaming output measurements. Then, a nonlinear MPC controller can be setup
by solving a finite-time nonlinear optimal control problem over the horizon $[k,k+p]$
at each execution time $k$ for a given prediction horizon $p$, taking $\hat x(k|k)$ as the initial state of the prediction and using $\hat\theta(k|k)$ to make nonlinear
predictions. An example of such an adaptive nonlinear MPC scheme,
that we will use in the numerical experiments reported in Section~\ref{sec:experiments}, is
\begin{equation}
    \ba{ll}
    \stackbin[u_0,\ldots,u_{p-1}]{}{\min}& \displaystyle{\sum_{t=0}^{p}}
    \|W^{\Delta u}(u_t-u_{t-1})\|_2^2+ \|W^{y}(y_{t}-r_{t})\|_2^2\\ 
    \hspace*{.5em}\st & \hspace*{-2.5em}x_{t+1}=f_x(x_t,u_t,\hat\theta_x(k|k)),\
        y_{t}=f_y(x_t,u_t,\hat\theta_y(k|k))\\
        &\hspace*{-2.5em}\umin\leq u_t\leq \umax\\
    \ea
\label{eq:NLMPC}
\end{equation}
where $x_0=x(k|k)$ and $u_{-1}=u(k-1)$. 
Note that the last penalty on $u_p-u_{p-1}$ and the first penalty on $y_0-r_0$
in~\eqref{eq:NLMPC} can be omitted in case of strictly-causal RNN models.

When full adaptation is not recommended, for instance to prevent excessive changes of the model parameters and/or to avoid computing the full EKF iterations, a non-adaptive nonlinear MPC setting
can be used, in which an optimal vector $\theta^*$
of parameters is obtained offline by running~\eqref{eq:EKF-full} on a 
training input/output dataset and used to make model-based predictions. 
A possible drawback of the latter approach is that offsets may arise in steady state
when tracking constant set-points
due to model/plant mismatches. A common practice is to augment the observer with a disturbance 
model, that in the current nonlinear MPC setting corresponds to augment 
the prediction model as follows (cf.~\cite{VBPP21}):
\begin{equation}
\ba{rcl}
    x(k+1)&=&f_x(x(k),u(k),\theta_x(k))+B_dd(k)+\xi(k)\\[.5em]    y(k)&=&f_y(x(k),u(k),\theta_y(k))+C_dd(k)+\zeta(k)\\[.5em]
    d(k+1)&=&d(k)+\eta(k)
\ea
\label{eq:disturbance-model}
\end{equation}
and estimate $\hat x(k|k)$, $d(k|k)$ by EKF. Clearly, such a solution
corresponds to only updating the bias terms $b_1^x$, $b_{L_y}^y$
of the RNN model~\eqref{eq:RNN} by setting $b_1^x(k|k)=(b_1^x)^*+B_dd(k|k)$,
$b_{L_y}^y(k|k)=(b_{L_y}^y)^*+C_dd(k|k)$. The (frequently used) case of
pure output disturbance models ($B_d=0$, $C_d=I$) corresponds
to only updating  $b_{L_y}^y$. 

\section{Numerical experiments}
\label{sec:experiments}
We test the proposed EKF-based RNN learning method on system
identification and nonlinear MPC problems. All computations are done
in MATLAB R2022b on an Apple M1 Max machine using CasADi~\cite{AGHRD19} for automatic differentiation.
We use the particle swarm optimizer PSwarm~\cite{VV09}
to solve the low-dimensional nonlinear programming problem~\eqref{eq:min-loss-state} 
with $\bar N=100$, initial population of $2n_x$ samples, and each component of $x_0$ 
restricted in $[-3,3]$. 

Unless stated differently, we use covariance matrices\footnote{We observed
that larger values of $Q_x(k)$, $Q_\theta(k)$ only slow down convergence speed without
providing any other benefit.}, $Q_x(k)\equiv 10^{-10}I$, $Q_\theta(k)\equiv 10^{-10}I$, $Q_y(k)\equiv 1$, 
$P(0|-1)$ as in~\eqref{eq:P0-1}, 
initial state $x(0|-1)=0$, $\gamma=10^{-4}$ in~\eqref{eq:min-loss-relaxed},
and Adam~\cite{KB14} over $N_e=500$ epochs 
with hyperparameters $\beta_1=0.9$, $\beta_2=0.999$ for SGD. 
Standard scaling of the input and output samples is performed by computing their means and standard deviations on training data. 
Model quality is judged in terms of the best fit rate (BFR) 
$100\left(1-\frac{\|Y-\hat Y\|_2}{\|Y - \bar y\|_2}\right)$
for numeric outputs, where $Y$ is the vector of measured output samples, $\hat Y$ the vector of
output samples simulated by the identified model fed in open-loop with the input data,
and $\bar y$ is the mean of $Y$, and by the accuracy
$
    a=\frac{1}{N}\sum_{k=1}^N \delta_{\hat y_b(k),y(k)}
$
for binary outputs, where $\hat y_b(k)=0$ if $\hat y(k)<0.5$
or $1$ otherwise. 
In all tests, we initialize the weights of the neural networks of the model
by using Xavier initialization~\cite{GB10}, with zero bias terms. 

\subsection{Fluid damper benchmark}
\label{sec:fluiddamper}
We consider the magneto-rheological fluid damper
problem~\cite{WSCH09} used in the System Identification (SYS-ID) Toolbox for MATLAB R2022b~\cite{Lju01} for nonlinear autoregressive (NARX) model identification, which
consists of $N=2000$ training data and $1499$ test data.
We want to train a RNN model~\eqref{eq:RNN} with $n_x=4$ hidden states and
shallow state-update and output network functions ($L_x=L_y=2$) with
$n_1^x=n_1^y=6$ neurons, arctangent activation functions $f_1^x,f_1^y$, 
and linear output function $f_2^y$, and $\ell_2$-regularization penalties
$\rho_\theta=\rho_x=10^{-3}$.
\begin{figure}[t]
\includegraphics[width=\hsize]{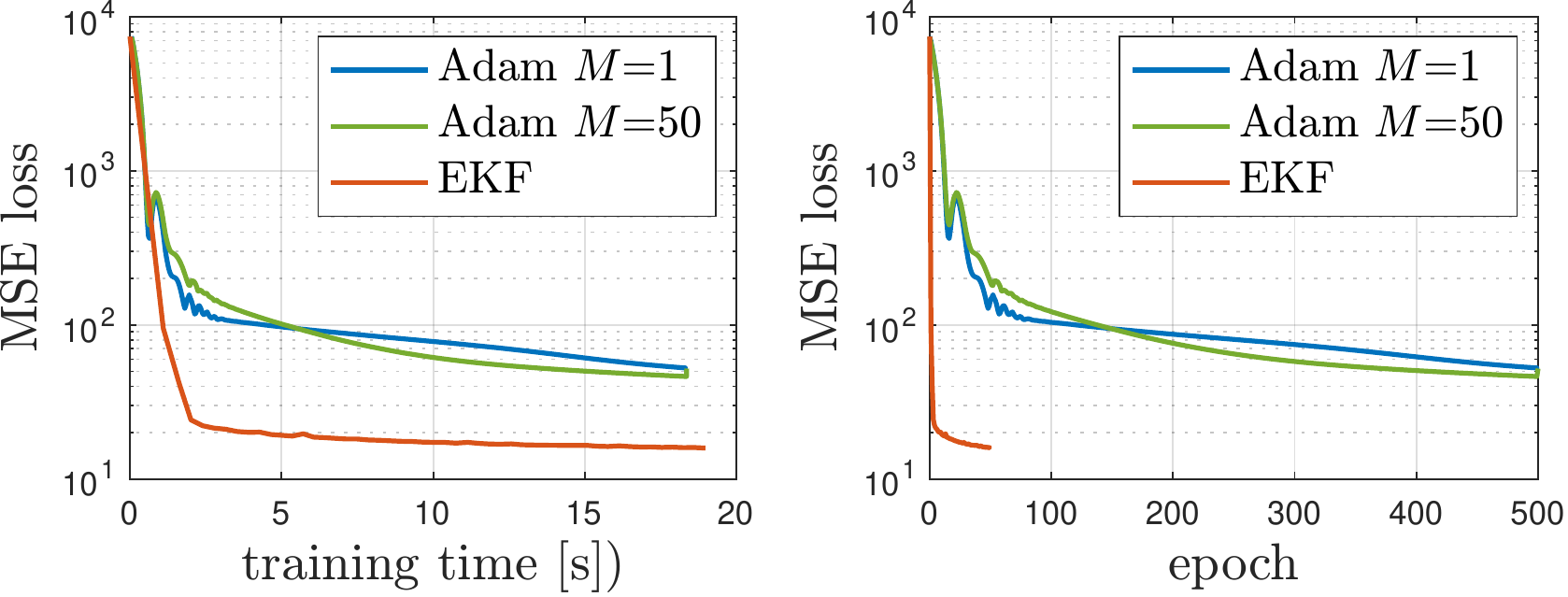}
\caption{Fluid damper benchmark: MSE loss $\frac{1}{2N}\sum_{k=0}^{N-1}(y(k)-\hat y(k))^2$
evaluated at each epoch}
\label{fig:V_train_fluiddamper}
\end{figure}
We run the EKF-based learning method over $N_e=50$ epochs and compare the results to
those obtained by solving the fully
condensed problem~\eqref{eq:min-loss} and the partially-condensed problem~\eqref{eq:min-loss-partial} with $M=50$ by using Adam with learning rate $l_r=0.005$ (the value of $l_r$ was chosen to get a good tradeoff between convergence speed and avoiding excessive oscillations).

Figure~\ref{fig:V_train_fluiddamper} shows the resulting MSE loss values in a typical run.
When using EKF, after each epoch the initial condition $x_0$ is reconstructed by solving~\eqref{eq:min-loss-state} with $\bar u(k)=u(k)$, $\bar y(k)=y(k)$ to
evaluate the MSE loss. 
It is apparent
that EKF reaches a good-quality model already after one pass through the training dataset
and outperforms the other methods.
Another advantage of EKF is that little effort was put on
tuning the covariance matrices $Q_y$, $Q_x$, $Q_\theta$, $P(0|-1)$
for the EKF, while Adam required a careful tuning of the learning rate $l_r$
and also of the penalty $\gamma$ in case $M>1$. 

Table~\ref{tab:fluiddamper_comparison} compares the fit results
obtained by running the training methods on both the same RNN model structure and 
an LSTM model~\cite{BTFS20} 
with 4 hidden and 4 cell states (i.e., $n_x=8$ states) 
and the same output function $f_y$. The table shows the mean and standard deviation 
obtained over 20 runs (with $N_e=25$ used for EKF), starting from different initial model parameters, always computing the fit on the model with the lowest MSE
obtained among all epochs. For further comparison, 
the best model \texttt{Narx\_6\_2} reported in~\cite{Lju01} 
provides a fit of 88.18\% on training data and 85.15\% on test data.

\begin{table}[h!]
\caption{Fluid damper benchmark: mean BFR (standard deviation) obtained over 20 runs}
\label{tab:fluiddamper_comparison}
\begin{center}
\begin{tabular}{l|l|c|c|c}
     &BFR    & Adam & Adam  & EKF \\
     &         & $M=1$ & $M=50$ & \\\hline
RNN    &training &  89.12 (1.83) & 88.56 (1.85)  & 92.82 (0.33)\\
$n_\theta=107$ &test &  85.51 (2.89) & 83.75 (4.71)  & 89.78 (0.58)\\\hline
LSTM   &training &  89.60 (1.34) & 87.47 (2.90)  & 92.63 (0.43)\\
$n_\theta=139$ &test &  85.56 (2.68) & 80.62 (6.89)  & 88.97 (1.31)\\\hline
\end{tabular}
\end{center}
\end{table}

\begin{figure}[t]
\includegraphics[width=\hsize]{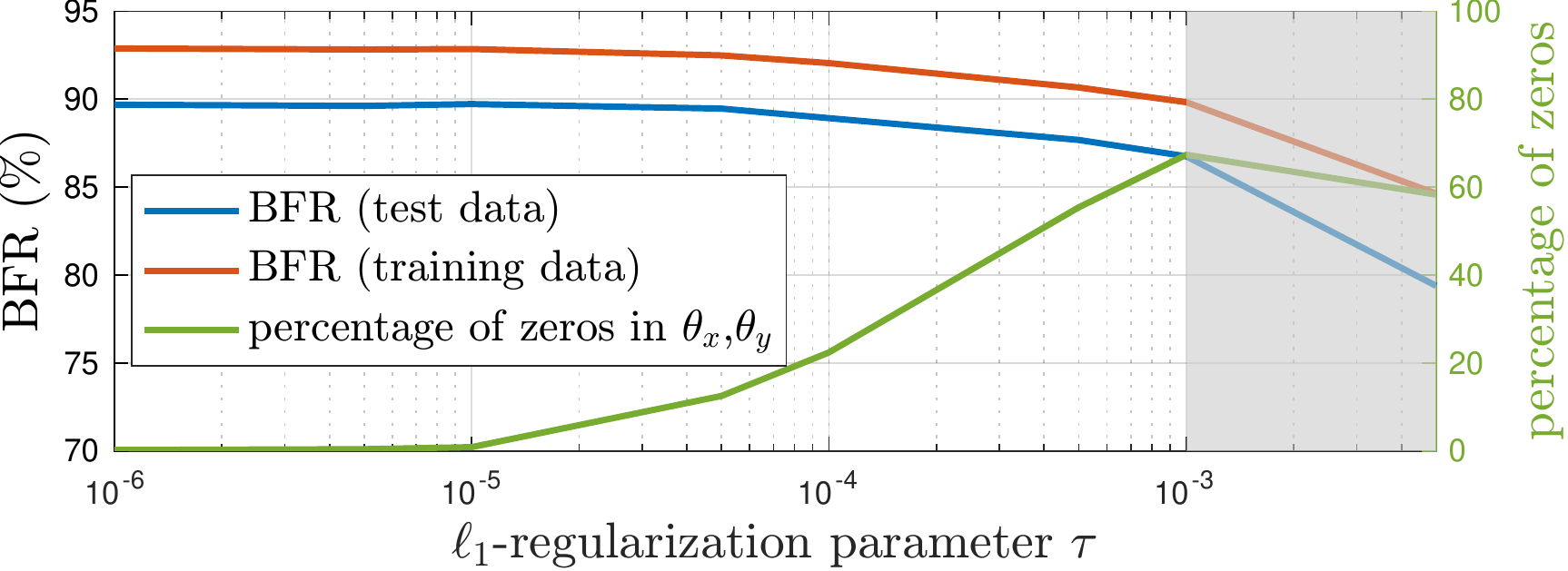}
\caption{Fluid damper benchmark: BRF and sparsity of $\theta$ 
optimized by EKF vs $\ell_1$-regularization coefficient $\lambda$
(mean values over 20 runs)}
\label{fig:L1_fluiddamper}
\end{figure}
Figure~\ref{fig:L1_fluiddamper} shows the mean BRF obtained 
over 20 runs when training the RNN model~\eqref{eq:RNN}
under $\ell_1$-regularization $\lambda \|\theta\|_1$, 
introduced in the EKF as in~\eqref{eq:l1-theta-update-2},
for different values of $\lambda$. The figure also shows the mean percentage of zero entries
in the resulting parameter vector $\theta$, where each entry $\theta_i$ such that $|\theta_i|\leq10^{-3}$
is set to zero after EKF training. 
As expected, for increasing values of $\lambda$ the parameter vector gets
more sparse, at the price of decreased prediction quality. For large values of $\lambda$
(roughly $\lambda>10^{-3}$), results start deteriorating, possibly due to the
excessively large steps taken in~\eqref{eq:l1-theta-update-2} that mine the convergence 
of the EKF.

\subsection{Linear dynamical system with binary outputs}
Consider $2000$ input/output pairs generated by the following linear system with binary outputs
\[
    \ba{rcl}
    x(k+1)&=&\smallmat{.8 & .2 & -.1\\ 0 & .9 & .1\\ .1 & -.1 & .7}x(k)+
             \smallmat{-1\\.5\\1}u(k)+\xi(k)\\[.5em]
    y(k)&=&\left\{\ba{ll}1&\mbox{if}\ \smallmat{-2 & 1.5 & 0.5}x(k)-{\scriptsize{2}}+\zeta(k)\geq 0\\0&\mbox{otherwise}\ea\right.
    \ea
\]
from $x(0)=0$, with the values of the input $u(k)$ changed with 90\% probability
from step $k$ to $k+1$ with a new value drawn from the uniform distribution on $[0,1]$. The disturbances $\xi_i(k),\zeta(k)\sim \NN(0,\sigma^2)$, $i=1,2,3$, are assumed independent. We consider the first $N=1000$ samples
for training, the rest for testing the model. We want to fit an affine model ($L_x=L_y=1$) with sigmoidal
output function $f_1^y(y)=1/(1+e^{-A_1^y[x'(k)\ u(k)]'-b_1^y})$. Output data
are not scaled.

We run EKF ($N_e=25$) and Adam with $\rho_x=\rho_\theta=10^{-2}$,
the initial weights randomly generated as in~\cite{GB10}
and further scaled by a factor $\frac{1}{20}$,
and the remaining settings as in Section~\ref{sec:fluiddamper},
modified cross-entropy loss $\ell_{\rm CE\epsilon}$ for $\epsilon=0.005$,
and EKF updates as in~\eqref{eq:CEloss}. Adam is run with 
learning rates selected by trial and error as $l_r=0.01$ ($l_r=0.001$) for $M=1$ ($M=50$)
to trade off convergence rate and variance. 
Table~\ref{tab:binary-outputs} shows the accuracy 
obtained by EKF and Adam (for $M=1$ and $M=50$)
for increasing values of $\sigma$. Note that we kept $Q_x=10^{-10}$ in all tests,
as we assumed not to know the intensity of the disturbances entering the system.

\begin{table}[h!]
\caption{Linear system with binary outputs: accuracy (\%) on test (training) data (mean values over 20 runs)}
\label{tab:binary-outputs}
\begin{center}
{\footnotesize
\begin{tabular}{c|c|c|c}
$\sigma$ & $M=1$  & $M=50$ & EKF\\[.5em]\hline
$0.000$&$97.37$ ($96.86$)&$83.06$ ($86.07$)&$98.02$ ($97.91$)\\
$0.001$&$95.00$ ($98.41$)&$86.88$ ($88.41$)&$95.33$ ($98.66$)\\
$0.010$&$97.38$ ($97.47$)&$87.65$ ($85.59$)&$97.99$ ($98.52$)\\
$0.100$&$94.84$ ($94.49$)&$74.64$ ($83.94$)&$94.56$ ($95.44$)\\
$0.200$&$91.49$ ($90.80$)&$80.37$ ($82.88$)&$93.71$ ($92.22$)\\
\hline
\end{tabular}
}
\end{center}
\end{table}

\subsection{Nonlinear MPC benchmark: ethylene oxidation plant}
We consider data generated from the ethylene oxidation plant model 
used as a nonlinear MPC benchmark in the Model Predictive Control Toolbox for MATLAB. A dataset of $2000$ samples is generated by numerically
integrating the system of nonlinear ordinary differential equations of the plant
model with high accuracy and collecting samples every $T_s=5$~s. 
The plant model has 4 states
(gas density, C2H4 concentration, C2H4O concentration, and temperature in the reactor),
one output ($y$ = C2H4O concentration), and two inputs ($u$ = total volumetric feed flow rate, that can be manipulated, and $v$ = C2H4 concentration of the feed, which is a measured disturbance). Half the dataset ($N=1000$ samples) is used to train a RNN with $L_x=3$ (i.e., a two-layer state-update neural network), $L_y=1$, $n_x=4$, $n_1^x=6$, $n_2^x=4$, affine output function, sigmoidal activation functions $f_1^x,f_2^x$, and unit output function $f_1^y$. We run the EKF-based training algorithm
by processing the dataset in $N_e=20$ epochs, which takes 4.54~s on the target machine.
The resulting BRF is 94.33\% on training data and 89.08\% on test data.

\begin{figure}[t]
\centerline{\includegraphics[width=\hsize]{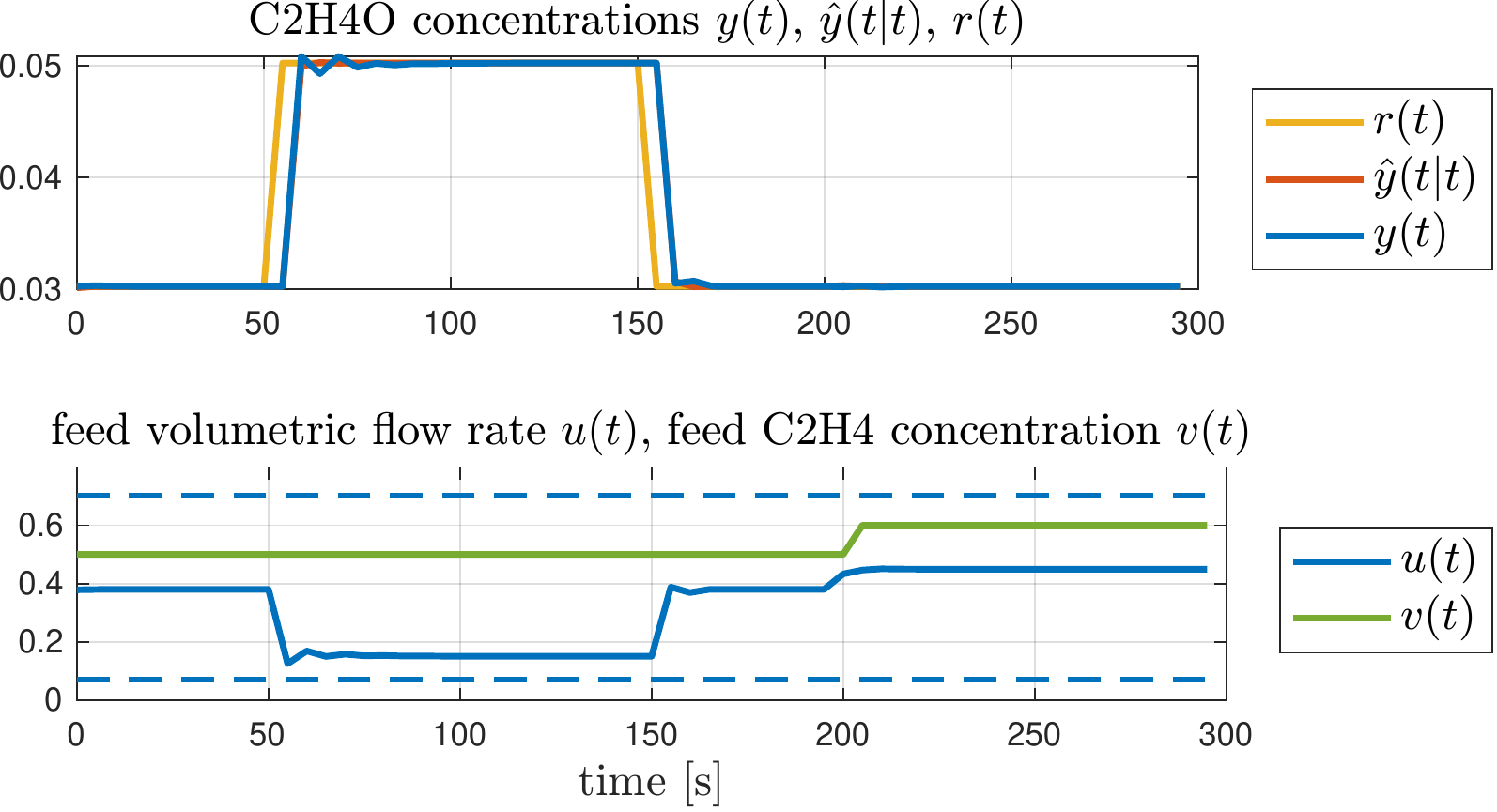}}
\caption{Ethilene oxidation benchmark: NLMPC results}
\label{fig:ethilene_NLMPC}
\end{figure}

The corresponding NLMPC controller, with MPC weights $W^{\Delta u}=0.1$, $W^y=10$, 
$\umin=0.0704$, $\umax=0.7042$, and prediction horizon $p=10$, is implemented using MATLAB's \texttt{fmincon} solver with default parameters and no Jacobian information, warm-started from the shifted previous optimal solution. To close the feedback loop and get offset-free tracking
of constant set-points, we apply EKF on line to estimate the state of 
the extended RNN model~\eqref{eq:disturbance-model}
with $B_d=0$, $C_d=1$ (output disturbance model); this corresponds
to only adapting the bias coefficient $b_1^y$ in~\eqref{eq:RNN-y},
with covariances $E[\xi(k)\xi(k)']=0.01 I$, $E[\eta(k)^2]=1$,
$E[\zeta(k)^2]=0.01$, and $P(0|-1)=I$. The obtained closed-loop results are depicted in Figure~\ref{fig:ethilene_NLMPC},
where it is apparent that a very good tracking is achieved despite the black-box
RNN model used for prediction. The execution time ranges between 
1.35~ms and 35.15~ms (8.86~ms on average) to solve the NLMPC problem
and between 0.03~ms and 4.58~ms (0.15~ms on average) for state estimation.

\section{Conclusions}\label{sec:conclusions}
We have shown that EKF is an effective way of learning control-oriented RNN models 
from input and output data, even in the case of general strongly convex and 
twice-differentiable loss functions and $\ell_1$-regularization. The approach is
particularly suitable for online learning and model adaptation
of recurrent neural networks
and can be immediately extended to handle other classes of parametric 
nonlinear state-space models.

An interesting topic for future research
is to study the conditions to impose on the RNN structure to make
$x$ and $\theta$ observable, in particular to prevent over-parameterizing 
the model, and on how to choose model structure, loss function, and regularization terms
to guarantee the asymptotic convergence of the filter.

\end{document}